\pdfoutput=1
\documentclass[11pt]{article}

\usepackage{ACL2023}

\usepackage{times}
\usepackage{latexsym}

\usepackage[T1]{fontenc}
\usepackage[utf8]{inputenc}

\usepackage{booktabs}
\usepackage{graphicx}
\usepackage{multirow}
\usepackage{amssymb}
\usepackage{makecell}

\usepackage{pifont}% http://ctan.org/pkg/pifont
\usepackage{colortbl}
\usepackage{amsmath} 
\usepackage{xspace}
\definecolor{mylightgreen}{RGB}{144,238,144}
\definecolor{mylightred}{RGB}{255,182,193}
\definecolor{mydarkgreen}{RGB}{0, 150, 0}

\usepackage{microtype}

\usepackage{inconsolata}

\newcommand{\kd}{\texttt{KD}\xspace}
\newcommand{\nokd}{\texttt{No-KD}\xspace}

\title{Knowledge Distillation vs. Pretraining from Scratch \\ under a Fixed (Computation) Budget}

\author{Minh Duc Bui$^\nabla$\quad~ Fabian David Schmidt$^\diamondsuit$\quad~\\ \textbf{ Goran Glavaš$^\diamondsuit$\quad~ Katharina von der Wense$^{\nabla\spadesuit}$} \\   
$^\nabla$Johannes Gutenberg University Mainz, Germany \\  
$^\diamondsuit$Center For Artificial Intelligence and Data Science, University of Würzburg, Germany\\  
$^\spadesuit$University of Colorado Boulder, USA\\  
{\tt \{minhducbui, k.vonderwense\}@uni-mainz.de} \\
{\tt \{fabian.schmidt, goran.glavas\}@uni-wuerzburg.de}}

\begin{document}
\maketitle
\begin{abstract}
% Knowledge distillation (KD) traditionally involves a multi-step process during training, requiring an additional forward pass through a substantial larger teacher model. On the other hand, training without \kd ( \nokd) omits the teacher model, enabling a more efficient token throughput. Scaling laws have indicated that smaller models can attain comparable performance to larger models by processing more tokens during pretraining within a fixed compute budget \cite{kaplan2020scaling}. We begin by addressing the need for a fair comparison between models pretrained with \kd and with  \nokd, see Table \ref{tab:time_budget}. We proceed to investigate the logical follow-up question: Does \kd for pretraining remain a viable strategy when accounting for compute budget? In an optimal setting where \nokd has sufficient data without sample repetition, we observe comparable performance with a vanilla-\kd strategy pretrained for the same duration as  \nokd, improving by 0.4 and 0.1 on average for 6- and 12-layer models on GLUE. On the other hand, more sophisticated strategies such as TinyBERT \cite{jiao-etal-2020-tinybert} consistently outperform standard pretraining by a significant margin on GLUE. Further analysis demonstrates that in scenarios requiring data repetition, KD's superior sample efficiency becomes even more advantageous compared to  \nokd, which, despite its speed, demands more epochs.%, whereas \kd extract more information from limited data.
Compared to standard language model (LM) pretraining (i.e., from scratch), Knowledge Distillation (\kd) %in language modelling 
entails an additional forward pass through a teacher model that is typically substantially larger than the target student model.
As such, \kd in LM pretraining materially 
slows down throughput of pretraining instances vis-a-vis pretraining from scratch. % (\nokd)
Scaling laws of LM pretraining suggest that smaller models can close the gap to larger counterparts if trained on more data (i.e., processing more tokens)---and under a fixed computation budget, smaller models \textit{are able} be process more data than larger models.  %during pretraining within a fixed compute budget.
%Thus, it remains unclear whether 
We thus \textit{hypothesize that \kd 
%for LM pretraining 
might, in fact, be suboptimal to pretraining from scratch for obtaining smaller LMs, when appropriately accounting for the compute budget}. To test this, we compare pretraining from scratch against several \kd strategies for masked language modeling (MLM)
%in encoder-only models 
in a \textit{fair} experimental setup, with respect to amount of computation as well as pretraining data.
%We observe that, 
Downstream results on GLUE, however, do \textit{not} confirm our hypothesis: while pretraining from scratch performs comparably to ordinary \kd under a fixed computation budget, 
%when unlimited pretraining tokens are available within the fixed compute budget, 
more sophisticated \kd strategies, namely TinyBERT \cite{jiao-etal-2020-tinybert} and MiniLM \cite{wang-etal-2023-distill}, outperform it by a notable margin. We further find that \kd yields larger gains over pretraining from scratch when the data must be repeated under the fixed computation budget.\footnote{Code is available at \url{https://github.com/MinhDucBui/revisiting_distillation}.}

\end{abstract}

\section{Introduction}

Knowledge distillation \citep[\kd;][]{hinton2015distilling,jiao-etal-2020-tinybert} during LM pretraining has emerged as the primary mean of compressing the capabilities of a large pretrained teacher model into a task agnostic smaller student model. \kd is praised for yielding high-performing task agnostic small models, mitigating the need for pretraining (small models) from scratch, which is typically considered more expensive. %\kd is touted for yielding performant small models, mitigating the need for pretraining (small models) from scratch, typically considered more expensive. 
The body of existing \kd work for MLM \cite{jiao-etal-2020-tinybert, wang-etal-2023-distill}, however, typically does not compare \kd against pretraining from scratch in a \textit{fair setup}: (i) with the same target models (exactly the same architecture) and (ii) under the same computation budget. Compared to just training the target model from scratch, \kd comes with a computational overhead of forward passes through the typically considerably larger teacher model. This, under the same computation budget, allows pretraining from scratch to consume more data (i.e., more tokens) than \kd, which leads to the central research question of this work: \textit{in a fair setup where both are given equal overall computation budget, is \kd still more effective than pretraining from scratch (\nokd)?} We hypothesize that, under a fair evaluation setup, \nokd may be as effective as \kd, rendering \kd inconsequential. Our reasoning is based on two observations:

\begin{table}
  \centering\small\setlength{\tabcolsep}{3pt}

  \begin{tabular}{lcc}
    \toprule
    & \multicolumn{2}{c}{\textbf{Identical}}  \\
    \cline{2-3}
    \textbf{Name} & \textbf{Architect.} & \textbf{Compute}\\
    \midrule
    DistilBERT  \cite{sanh2020distilbert} & \textbf{\textit{No}} & \textbf{\textit{No}} \\
    TinyBERT \cite{jiao-etal-2020-tinybert} & \textbf{\textit{Yes}} & \textbf{\textit{No}}  \\
    MobileBERT \cite{sun-etal-2020-mobilebert} & \textbf{\textit{No}} & \textbf{\textit{No}} \\
    MiniLM \cite{wang2020minilm} & \textbf{\textit{No}} & \textbf{\textit{No}} \\
    Our Work & \textbf{\textit{Yes}} & \textbf{\textit{Yes}} \\
    \bottomrule
  \end{tabular}
  \caption{Assessing the fairness of evaluation setups in previous works for task-agnostic masked language models, trained with \kd and without \kd.}
  \label{tab:time_budget}
\end{table}

\paragraph{1) Fair \kd Comparison.} %First, \kd typically entails a multi-step procedure. The training instance is first passed through a large teacher model. This additional step typically slows down training significantly. %, but is not acknowledged by existing evaluation protocols, which ignore the overall computation budget of \kd. 
%Then, a student model is trained on both the original task and matching the output distribution of the teacher model.
% This process includes a forward pass through a large teacher model, followed by a forward pass through a smaller student model. %Subsequently, it involves a backward pass through the student model. 
%Compared to \kd, pretraining the target model from scratch (\nokd) eliminates the teacher model from the equation, allowing the training pipeline to operate more efficiently. This means that, under a fixed computation budget, LM training from scratch can process \textit{more tokens} than \kd. Nonetheless,
A fair comparison, in which both setups are given identical computation budgets (as well as identical target models) eludes existing work on \kd. %Extensive efforts have been dedicated to identifying optimal distillation strategies, ranging from distilling information directly from teacher logits to distilling intermediate representations like hidden states or attention blocks \citep{sanh2020distilbert, jiao-etal-2020-tinybert, sun-etal-2020-mobilebert, wang2020minilm}. 
\citet{jiao-etal-2020-tinybert} compare their model to BERT\textsubscript{Tiny} \cite{turc2019wellread}, which has the same architecture but employs significantly different training resources than their TinyBERT\textsubscript{Tiny}, preventing a fair comparison. Similarly, \citet{sanh2020distilbert} compare their distilled student solely against the teacher, whereas \citet{sun-etal-2020-mobilebert, wang2020minilm} only add comparison against larger pretrained models and competing \kd strategies. Even the body of work that focuses on comparing different \kd strategies has only recently sought to standardize training and thus enable fair comparisons \cite{lu2022knowledge,wang-etal-2023-distill}.
%Addressing this gap, our study conducts a fair comparison between encoder-only models trained using \nokd and \kd, considering compute budget, see Table \ref{tab:time_budget}.
%Notably, \citet{jha2023train} found that decoder-only large LMs perform similarly with \nokd and Vanilla-\kd on zero-shot tasks, and, unexpectedly, surpassing even more advanced \kd strategies. However, their study rather indicates that conventional \kd is less effective for GPT-style models. 
%Notably, \citet{jha2023train} show that decoder-only large LMs, when assessed on zero-shot tasks, reveal on par performance between \nokd and Vanilla-\kd, and even outperforms more sophisticatd \kd strategies. % However, their study rather indicates that conventional \kd is less effective for GPT-style models.

\paragraph{2) Scaling Laws.} Scaling laws \cite{kaplan2020scaling, hoffmann2022training}, reveal that, under a fixed computation budget, only a marginal correlation exists between the LM size and it's performance: 
%for a constant compute budget and data size, an optimal model size can be determined. Interestingly, 
smaller models compensate their lower learning efficiency with the ability to process more tokens within the same budget. While there are ongoing refinements to the scaling law \cite{hoffmann2022training}, it has been consistently reaffirmed by several studies \cite{geiping2022cramming, bansal2022data, clark2022unified}. 
%%%
For instance, \citet{geiping2022cramming} showcases this behavior by training multiple BERT models with varying architecture sizes for a fix 24 hour duration, resulting in similar loss values across all sizes. 

%Our main argument asserts that  \nokd, by virtue of its exclusion of the additional teacher inference, processes more tokens compared to \kd and may result in the same training efficiency achieved through \kd, thereby \textit{achieving a compensatory mechanism} analogous to what is observed in the domain of scaling laws.

\paragraph{Contribution} Motivated by the recent findings in the realm of scaling laws and recognizing the absence of a fair comparison between \kd and \nokd, our primary contribution lies in the comparison of \nokd against \kd strategies for MLM while ensuring a fair setup with regards to compute budget and pretraining data. We initially assess \nokd in an optimal setup, where unlimited pretraining tokens are available within a fixed compute budget. Additionally, we examine the scenario when data is constrained within a fixed compute budget.

Our results reveal that, in the optimal setting, \nokd performs indeed comparably to vanilla-\kd, exhibiting an average improvement over vanilla-\kd of 0.4 and 0.1 points for 6-, and 12-layer models on GLUE. However, \nokd falls short of surpassing more advanced \kd strategies, exemplified by the comparison with TinyBERT and MiniLM. When available data is limited within the fixed compute budget, \kd strategies outperform \nokd by an even larger margin: \nokd, though faster, needs more epochs, whereas \kd strategies extract more information from limited data.

%For simplicity, in this study we focus on task-agnostic encoder student models, obtained via masked language modeling (MLM) \cite{sanh2020distilbert}. We pose the following hypothesis: \textit{Conventional pretraining (\nokd) should yield similar results compared to employing \kd during pretraining if we consider a fair setting.} In our work, `fairness' requires that both models have an equal number of parameters \textit{and} the same compute 

\begin{table*}
    \centering
    \small
    \setlength{\tabcolsep}{6pt}
    \begin{tabular}{l|c|ccccccc|c|c}
        \toprule
        & \textbf{Total Token} & \textbf{QNLI} & \textbf{SST-2} & \textbf{MNLI} & \textbf{MRPC} & \textbf{QQP} & \textbf{RTE} & \textbf{CoLA} & \textbf{Avg} & $\mathbf{\Delta}$ \\
        & \textbf{Throughput} & (Acc) & (Acc) & (Acc) & (F1) & (Acc) & (Acc) & (Mcc) &  \\
        
        \midrule
        \multicolumn{10}{c}{\textit{6-Layer: Unlimited Pretraining Tokens within Fixed Compute Budget}} \\ \midrule
        \rowcolor{gray!20} \texttt{No-KD\textsubscript{6}} & 4.6B & 86.5 & 89.8 & 79.2 & 87.3 & 90.1 & 60.7 & \textbf{47.3} & 77.3 & -- \\
        
        \texttt{Vanilla-KD\textsubscript{6}} & 2.6B & 87.3 & 89.2 & 78.8 & 87.1 & 89.7 & 61.7 & 44.8 & 76.9 & \textcolor{mydarkgreen}{$-0.4$} \\
        \texttt{MiniLM\textsubscript{6}} & 2.6B & 88.5 & 90.6 & 81.7 & 90.0 & 90.3 & 64.3 & 43.9 & 78.5 & \textcolor{red}{$+1.2$} \\
        \texttt{TinyBERT\textsubscript{6}} & 2.6B & \textbf{89.5}  & \textbf{91.0}  & \textbf{82.2} & \textbf{90.3}  & \textbf{90.4} & \textbf{67.2} & 40.8  & \textbf{78.8} & \textcolor{red}{$+1.5$ }\\ 

        \midrule
        \multicolumn{10}{c}{\textit{12-Layer: Unlimited Pretraining Tokens within Fixed Compute Budget}} \\ \midrule
        \rowcolor{gray!20} \texttt{No-KD\textsubscript{12}} & 4.6B & 87.8 & 90.1 & 80.6 & 86.5 & 90.3 & 60.3 & \textbf{51.1} & 78.1 & -- \\ 
        
        \texttt{Vanilla-KD\textsubscript{12}} & 2.6B & 86.3 & 90.5 & 79.8 & 88.8 & 89.9 & 62.1 & 48.6 & 78.0 & \textcolor{mydarkgreen}{$-0.1$} \\
        
        \texttt{MiniLM\textsubscript{12}} & 2.6B & \textbf{90.0} & 91.2 & \textbf{83.3} & 90.1 & \textbf{90.9} & \textbf{69.0} & 49.1 & \textbf{80.5} & \textbf{\textcolor{red}{$+1.4$}} \\
        
        \texttt{TinyBERT\textsubscript{12}} & 2.6B & 89.5 & \textbf{91.4} & 82.0 & \textbf{90.8} & 90.6 & 65.5 & 41.1 & 78.7 & \textbf{\textcolor{red}{$+0.6$}} \\
        
        \midrule
        \midrule
        \multicolumn{10}{c}{\textit{6-Layer: Limited Pretraining Tokens within Fixed (Increased) Compute Budget}} \\ \midrule
        \rowcolor{gray!20} \texttt{No-KD\textsubscript{6}} & 27.9B & 88.8 & 91.2 & 81.3 & 88.0 & 90.4 & 59.6 & 50.5 & 78.5 & -- \\ 
        
        \texttt{Vanilla-KD\textsubscript{6}} & 15.4B & 86.9 & 91.1 & 81.1 & 89.5 & 90.3 & 61.7 & \textbf{58.3} & 79.8 & \textcolor{red}{$+$1.3} \\ 
        \texttt{MiniLM\textsubscript{6}} & 15.6B & 90.0 & 91.5 & 83.0 & \textbf{90.3} & 90.6 & 65.7 & 50.7 & 80.3 & \textcolor{red}{$+$1.8}\\

        \texttt{TinyBERT\textsubscript{6}} & 15.6B & \textbf{90.5} & \textbf{92.3} & \textbf{83.3} & 90.2 & \textbf{90.8} & \textbf{67.5} & 51.8 & \textbf{80.9} & \textcolor{red}{$+$2.4}\\ 
        
        \bottomrule
    \end{tabular}
    \caption{Upper part: optimal scenario for  \nokd~-- unlimited pretraining tokens within a fixed compute budget. Lower part: limited data within a fixed compute budget. We present the performance results on the GLUE development set, maintaining a consistent pretraining wall-clock time across all models within each group. The column \textbf{Avg} represents the average performance across all tasks, while $\mathbf{\Delta}$ quantifies the average difference between \texttt{No-KD\textsubscript{xx}} and the other distillation strategies.}
    \label{tab:main}
\end{table*}

\section{Distillation Strategies} \label{sec:strategy}

\paragraph{\texttt{Vanilla-KD}} 
% The conventional approach to distill knowledge from a large teacher involves mimicking the teacher's predictions by minimizing the soft cross-entropy between the teacher's logits $\mathbf{z}^T$ and the student's logits $\mathbf{z}^S$, with a temperature factor $t$: $\mathcal{L}_{\text{pred}} = \texttt{CE}(\mathbf{z}^T/t, \mathbf{z}^S/t)$. Following \citet{hinton2015distilling}, the final training loss combines $\mathcal{L}_{\text{pred}}$ with the masked language modeling loss $\mathcal{L}_{\text{CE}}$ during pretraining, which we refer to as \texttt{Vanilla-KD}. 
Vanilla-\kd for MLM pretraining is set up as follows. 
A small MLM student is trained to mimick the predictions for a particular training instance of a large pretrained MLM teacher: The distillation objective is to minimize the soft cross-entropy between the logits $\mathbf{z}^T$ of the MLM teacher and the logits $\mathbf{z}^S$ of the MLM student, with a temperature factor $t$: $\mathcal{L}_{\text{pred}} = \texttt{CE}(\mathbf{z}^T/t, \mathbf{z}^S/t)$. Following \citet{hinton2015distilling}, the final training loss equally combines $\mathcal{L}_{\text{pred}}$ with the MLM loss $\mathcal{L}_{\text{CE}}$ during pretraining. %We refer to this as \texttt{Vanilla-KD}. 

% \paragraph{\texttt{TinyBERT}} \citet{jiao-etal-2020-tinybert}  hidden states and multi-head attention blocks of specific layers from the teacher. To achieve this objective, they utilize the following loss functions: $\mathcal{L}_{\text{hid}} = \texttt{MSE}(\mathbf{H}^S\mathbf{W_h}, \mathbf{H}^T)$ for the hidden states, and $\mathcal{L}_{\text{att}} = 1/h \sum_{i=1}^h \texttt{MSE}(\mathbf{A}_i^S, \mathbf{A}_i^T)$ for the multi-head attention matrices. Here, $\mathbf{H}^S$ and $\mathbf{H}^T$ represent the hidden states, and $\mathbf{A}_i^S$, or $\mathbf{A}_i^T$ denotes the \textit{unnormalized} attention value matrix of the $i$-th head corresponding to the teacher or student, respectively. Finally, embedding distillation is applied, where $\mathcal{L}_{\text{embd}} = \texttt{MSE}(\mathbf{E}^S \mathbf{W}_e, \mathbf{E}^T)$, with $E^S$ and $E^T$ denoting the embeddings of the student and teacher networks, respectively. The matrices $\mathbf{W}_h$ and $\mathbf{W}_e$ represent learnable projection matrices designed to distill across various sizes of architectures.
\paragraph{\texttt{TinyBERT}} 
%In addition to \texttt{Vanilla}-\kd, \citet{jiao-etal-2020-tinybert} further layer-wise distill the embeddings, the attention blocks, and the hidden states of transformer layers from the teacher LM to the student LM. 
\citet{jiao-etal-2020-tinybert} distill knowledge by minimizing the mean-squared error (\texttt{MSE}) between latent representations of the MLM student $S$ and the MLM teacher $T$ by model layers as follows. First, the embedding matrices of the student ($\mathbf{E_S}$) and the teacher ($\mathbf{E_T}$) are aligned by minimizing the loss $\mathcal{L}_{\text{embd}} = \texttt{MSE}(\mathbf{E}^S \mathbf{W}_e, \mathbf{E}^T)$. The authors further fit the unnormalized attention scores per head $h$ of the MLM student $S$ to the MLM teacher $T$ by optimizing $\mathcal{L}_{\text{att}} = 1/h \sum_{i=1}^h \texttt{MSE}(\mathbf{A}_i^S, \mathbf{A}_i^T)$. Lastly, the output hidden states $\mathbf{H}^S$ of transformer layers of the student are also regressed onto the corresponding teacher output representations $\mathbf{H}^T$ by optimizing $\mathcal{L}_{\text{hid}} = \texttt{MSE}(\mathbf{H}^S\mathbf{W_h}, \mathbf{H}^T)$.\footnote{The distillation of embeddings $\mathbf{E}$ and output hidden states $\mathbf{H}$ is learned up to projection matrices $\mathbf{W}_{e;h}$ matrices to bridge varying dimensionalities of representations across architectures.}
% Lastly, the predictions of the student MLM are also trained to match the predictions of the teacher MLM (cf. \texttt{Vanilla}-\kd).
% TODO: maybe a sentence of layerwise alignment ie 6L student 12L teacher every second layer 

\paragraph{\texttt{MiniLM}} 

\citet{wang2020minilm} also mimic the self-attention modules of the MLM teacher. Unlike \texttt{TinyBert}, \texttt{MiniLM} focuses on the last attention module.  \citet{wang2020minilm} minimize the KL-divergence between the self-attention distributions of the MLM teacher and the MLM student. They further minimize the KL-divergence between the value relations of the MLM teacher $T$ and MLM student $S$, i.e. $\mathcal{L}_{VR} = \frac{1}{A_{h}|x|}\sum_{a=1}^{A_h}\sum_{t=1}^{|x|}D_{KL}(\mathbf{VR}^T||\mathbf{VR}^S)$. The value-relation denotes the outer product of values $\mathbf{V}$ across heads in the last attention module, i.e. $ \mathbf{VR} = softmax(\frac{\mathbf{V}\mathbf{V}^{\texttt{T}}}{\sqrt{d}})$. % The value relation is computed through the multi-head scaled dot-product between values. % Additional details can be found in \citet{wang2020minilm}.

\section{Experiment Setup}

\paragraph{Model Architectures} We experiment with two different teacher and student sizes: First, we use a 12-layer pretrained BERT\textsubscript{base} \cite{devlin-etal-2019-bert} model (L=12, H=768, A=12, Total Parameters=110M) as the teacher and a randomly initialized 6-layer BERT\textsubscript{6} model (L=6, H=768, A=12, Total Parameters=67M) as the student. We then scale the setting up to a pretrained BERT\textsubscript{Large} (L=24, H=1024, A=16, Total Parameters=340M) teacher and a randomly initialized 12-layer BERT\textsubscript{12} student. To speed up the training pipeline and convergence, we use the implementation of \citet{izsak-etal-2021-train} for the models.

\paragraph{Data} We follow BERT \cite{devlin-etal-2019-bert} and pretrain all models on the Toronto BooksCorpus \cite{zhu2015aligning} and English Wikipedia.\footnote{In November 2023, we crawled the English Wikipedia using \citet{Wikiextractor2015}. The official BookCorpus is no longer accessible; however, it was re-crawled by \citet{soskkobayashi2018bookcorpus}.} After MLM pretraining, we finetune and evaluate the models on the General Language Understanding Evaluation (GLUE) benchmark \cite{wang-etal-2018-glue}, a collection of diverse natural language understanding tasks.\footnote{We refer to Appendix \ref{sec:glue} for more information about the GLUE datasets.}

\paragraph{Pretraining} We first pretrain a BERT\textsubscript{6} model from scratch (without \kd). This model is denoted as \texttt{No-KD\textsubscript{6}}. We further apply the \kd strategies (cf. \S\ref{sec:strategy}), for which BERT\textsubscript{base} (BERT\textsubscript{large}) is the MLM teacher for the 6-layer (12-layer) MLM student. The resulting 6-layer models are indicated with a subscript 6, while the 12-layer models are marked with a subscript 12. Notably, we only employ the last layer for layer-wise distillation, as we confirm the findings of \citet{wang-etal-2023-distill} that distilling knowledge of multiple layers does not yield consistent performance improvements. We refer to Appendix \ref{sec:pretraining} for additional hyperparameter, hardware details and wall-clock training time.

%Our pretraining pipeline employs a batch size of 1024, employing gradient accumulation with a batch size of 256. We adopt a time-based learning rate schedule with a linear curve. The peak learning rate is set to 5e-4 for distillation strategies and 1e-3 for \nokd. We opt for a warmup proportion of 0.06 for both scenarios. Utilizing the AdamW optimizer with $(\beta_1, \beta_2) = (0.9, 0.98)$ and $\epsilon = 1e-6$, we conduct training with mixed precision techniques. We measure compute budget by wall-clock time.

%We use a batch size of 1024, a maximum sequence length of 128 and a peaking learning rate of 5e-4 for distillation strategies and 1e-3 for \texttt{No-KD}. We refer to Appendix \ref{sec:pretraining} for more information about the pretraining details.

%We train with two different training times: Initially, we conduct experiments by measuring the time it takes to pretrain \texttt{No-KD\textsubscript{6}} for one epoch on our English corpus. We then fix this time and proceed to train all models for that specified pretraining duration. Subsequently, we measure the time for 6 epochs and train all models for the resulting training duration.

\paragraph{Downstream Finetuning} We perform a grid search over batch sizes \{16, 32\} and learning rates \{1e-5, 3e-5, 5e-5, 8e-5\} to identify the ideal hyperparameters for each task on the GLUE benchmark. We train all configurations for 5 epochs. We utilize a polynomial learning rate schedule and a maximum sequence of 128. 

\section{Results}

\subsection{Setting: Unlimited Data with Fixed Compute}

We assess \nokd for a single epoch and fix the resulting training wall-clock time for the distillation strategies. Within this compute budget, we train on unlimited pretraining tokens without the need for sample repetition. We report our main results in the upper segment of Table \ref{tab:main}.

\paragraph{Low \kd Token Throughput}

% First, we present the token throughput across various models and settings.
We find that the token throughput of \texttt{No-KD\textsubscript{6}} and \texttt{No-KD\textsubscript{12}} is approximately 1.8 times greater than that of the distillation models. This observation underscores that the presence of a teacher model greatly reduces the speed of pretraining. 
% The following investigation revolves around whether the increased token throughput achieved with \nokd translates into a substantial performance boost.

\paragraph{Performance of 6-layer Students} We observe that \texttt{No-KD\textsubscript{6}} surpasses \texttt{Vanilla-KD\textsubscript{6}} by an average of 0.4 points.
% This result indicates that \texttt{Vanilla-KD\textsubscript{6}} is not superior to pretraining from scratch in a fair setting.
This result indicates that \texttt{Vanilla-KD\textsubscript{6}} does not exceed pretraining from scratch in a fair setting.
However, more advanced \kd strategies exhibit notable performance gains over \nokd. 
% For instance, \texttt{TinyBERT\textsubscript{6}} outperforms \texttt{No-KD\textsubscript{6}} by 1.5 average points, while \texttt{MiniLM\textsubscript{6}} achieves a 1.2 point advantage on average.
On average, \texttt{TinyBERT\textsubscript{6}} outperforms \texttt{No-KD\textsubscript{6}} by 1.5 points, while \texttt{MiniLM\textsubscript{6}} achieves a 1.2 point advantage.
These findings suggest that pretraining from scratch falls short in outperforming sophisticated distillation strategies in a fair setup, even when exposed to a higher volume of tokens. The only exception to this trend is CoLA (Corpus of Linguistic Acceptability) \cite{warstadt-etal-2019-neural}, on which \texttt{No-KD\textsubscript{6}} excels.

\paragraph{Performance of 12-layer Students} 
% We find the same pattern for the 12-layer model setup:
We find the same pattern when we double the number of transformer layers in MLM students:
\texttt{Vanilla-KD\textsubscript{12}} fails to outperform \texttt{No-KD\textsubscript{12}}, yet it is surpassed by 
%more involved strategies like
\texttt{MiniLM\textsubscript{12}} by an average of 1.4 points. Notably, \texttt{No-KD\textsubscript{12}} once again exhibits superior performance on the CoLA task compared to other strategies. %As an additional observation, \texttt{TinyBERT\textsubscript{12}} does not scale effectively when transitioning to a 12-layer student model under fixed data conditions, only increasing performance.

\paragraph{CoLA Performance} \texttt{No-KD\textsubscript{6}} demonstrates superior performance on CoLA, surpassing the next most effective strategy by 2.5 and 2.0 points for 6- and 12-layer models, respectively. We hypothesize that CoLA benefits significantly from masked language modelling, as evidenced by the improved performance of \texttt{Vanilla-KD} on CoLA compared to other distillation strategies, aligning with findings by \citet{wang-etal-2023-distill}. Another contributing factor could be the scalability of CoLA with respect to tokens encountered during pretraining. This observation contradicts the results of \citet{liu-etal-2021-probing-across}, who suggest that CoLA can be learned relatively quickly compared to other downstream tasks. However, it aligns with the conclusions of \citet{geiping2022cramming}, who also note that their BERT version, exposed to less data, exhibits subpar performance on CoLA.

\subsection{Setting: \textit{Limited} Data with Fixed Compute}

% To broaden the scope of our findings, we increase the compute budget while maintaining a consistent dataset size.
To extend our findings, we increase the compute budget while retaining a fixed dataset size. We evaluate this setup with 6-layer MLM students and the 12-layer MLM teacher.
% This results in a scenario where we are working with limited data within a fixed compute budget.
The analysis provides an estimate of the viability of \nokd when data repetition is necessary within the fixed compute budget. %Our hypothesis leans towards a pessimistic scenario, given that \nokd benefits from observing a greater number of tokens, demanding a larger dataset for effective scaling. While this demand may be met in high-resource languages with recent datasets \cite{kudugunta2023madlad400}, it poses a challenge in mid to low-resource language scenarios, which could potentially benefit the most from our hypothesis. 
The results are presented in the lower section of Table \ref{tab:main}.

% \paragraph{Result} 
The \texttt{No-KD\textsubscript{6}} model is underperforming, compared to all distillation strategies, including \texttt{Vanilla-KD\textsubscript{6}} by 1.3 points. The performance gap widens even more when compared to \texttt{MiniLM\textsubscript{6}} and \texttt{TinyBERT\textsubscript{6}}, with a substantial difference of 1.8 and 2.4 points on average. We attribute this to the fact that while \nokd benefits from exposure to a larger number of tokens, it also necessitates a larger dataset for effective scaling. Although this requirement can be met in high-resource languages with up-to-date datasets \cite{kudugunta2023madlad400}, it presents a significant challenge in mid to low-resource scenarios.
Additionally, \texttt{No-KD\textsubscript{6}} is now being outperformed even on CoLA. These results suggest that CoLA's performance indeed needs to process a certain quantity of tokens during pretaining to scale effectively, regardless of additional token repetitions: e.g., the performance of \texttt{Vanilla-KD\textsubscript{6}} increases by 13.5 points if scaled from 2.6B unique to 15.4B non-unique pretraining tokens. Interestingly, our findings reveal that \texttt{Vanilla-KD\textsubscript{6}} exhibits the best performance on CoLA, underscoring the advantageous impact of masked language modelling on this particular dataset.

%Our \texttt{No-KD\textsubscript{6}} model is outperformed already by the \texttt{Vanilla-KD\textsubscript{6}} strategy by 1.3 points on average. The margin between \texttt{No-KD\textsubscript{6}} and \texttt{TinyBERT\textsubscript{6}} is even larger with 2.4 points on average. We reason by that \nokd benefits from observing a greater number of tokens, it also demands a larger dataset for effective scaling. While this demand may be met in high-resource languages with recent datasets \cite{kudugunta2023madlad400}, it poses a challenge in mid to low-resource language scenarios.

%Now, even on CoLA \texttt{No-KD\textsubscript{6}} is getting outperformed. The results indicate that indeed CoLA scales well with the amount of tokens seen during training. We further see that \texttt{Vanilla-KD\textsubscript{6}} performs best on CoLA indicating that CoLA benefits from MLM.% when data is scarce (can do multiple epochs), \texttt{No-KD\textsubscript{6}} falls short by a large margin compared to distillation strategies. We hypothesize that because of seeing multiple samples more than once, the sample-efficiency decreases and therefore distillation strategies can increase the gap even more.

\section{Discussion}

While our study provides insights into a fair evaluation of \nokd and \kd for encoder-only models of moderate sizes, revealing negative results for \nokd, it may not cover the full spectrum of model sizes and architectures. For instance, \citet{jha2023train} show that for large decoder-only language models, \nokd performs comparably to \texttt{Vanilla-KD}, aligning with our findings. However, advanced \kd strategies like \texttt{MiniLM} exhibit poorer performance than \nokd and \texttt{Vanilla-KD}, challenging both our results and common beliefs about \kd regarding large decoder models. This disparity underscores the need for further investigation into a fair \kd evaluation across a range of \textit{architectures and scales}. Additionally, we recommend investigating the impact of the teacher budget on performance in the fair setting, a consideration not closely examined in our current work.% Nevertheless, our study provides supplementary insights into encoder-only models of moderate size.

\section{Conclusion}

In this work, we investigate our hypothesis that, provided a fair training scenario, model pretraining from scratch yields similar results as \kd during pretraining. Our rationale is grounded in recent advancements in scaling laws for language models and that the literature lacks a fair comparison between \nokd and \kd. Our findings demonstrate that our initial assumption does \textit{not} hold true: while, in an optimal setting for \nokd, \nokd performs on par with ordinary \kd, it falls short when compared to more sophisticated \kd strategies.

\section*{Limitations}
Firstly, we acknowledge that assessing the compute budget based on training wall-clock time comes with inherent limitations. As outline in \citet{kaddour2023train}, wall-clock time can fluctuate even on identical hardware. This fluctuation may arise from factors such as the utilization of non-deterministic operations or hidden background processes. Nevertheless, we only see negligible variations across different runs for the same training pipeline.

%Furthermore, our investigation is exclusively focused on encoder-only models of relatively moderate size. While our study provides insights into the fair evaluation of \nokd and \kd, it may not cover the full spectrum of model sizes and architectures. For instance, \citet{jha2023train} show that for large decoder-only language models, \nokd performs comparably to \texttt{Vanilla-KD}, aligning with our findings. However, advanced \kd strategies like \texttt{MiniLM} exhibit poorer performance, challenging both our results and common beliefs about \kd. This disparity underscores the need for further investigation into \kd strategies across a range of architectures and scales. Additionally, we recommend investigating the impact of teacher budget on performance in our fair comparison study, a consideration not closely examined in our current work. Nevertheless, our study provides supplementary insights into encoder-only models.

Another limitation of our work pertains to data size. Exploring larger pretraining corpora than ours might be worthwhile, although we note that even within our current data scale, \kd consistently outperforms \nokd by a significant margin. Even with potential increases in data size, \kd remains valuable as it provides a stronger starting point compared to \nokd.

Lastly, we acknowledge that the pretraining corpus is the same as what the teacher used. This shared corpus might influence \kd strategies either positively or negatively.

\section*{Ethics Statement}

We acknowledge our exclusive focus on the English language, overlooking the many challenges of other languages. Additionally, we recognize our sole emphasis on performance metrics, neglecting considerations related to the fairness of the resulting models. We also note that our research extensively employed GPU hours for both pretraining and finetuning, with a keen awareness of the environmental and resource implications associated with such usage. 

\section*{Acknowledgement}

The research in this paper was funded by the Carl Zeiss Foundation, grant number P2021-02-014 (TOPML project).

\bibliography{anthology,custom}

\begin{thebibliography}{29}
\expandafter\ifx\csname natexlab\endcsname\relax\def\natexlab#1{#1}\fi

\bibitem[{Attardi(2015)}]{Wikiextractor2015}
Giusepppe Attardi. 2015.
\newblock Wikiextractor.
\newblock \url{https://github.com/attardi/wikiextractor}.

\bibitem[{Bansal et~al.(2022)Bansal, Ghorbani, Garg, Zhang, Cherry, Neyshabur, and Firat}]{bansal2022data}
Yamini Bansal, Behrooz Ghorbani, Ankush Garg, Biao Zhang, Colin Cherry, Behnam Neyshabur, and Orhan Firat. 2022.
\newblock \href {https://proceedings.mlr.press/v162/bansal22b.html} {Data scaling laws in {NMT}: The effect of noise and architecture}.
\newblock In \emph{Proceedings of the 39th International Conference on Machine Learning}, volume 162 of \emph{Proceedings of Machine Learning Research}, pages 1466--1482. PMLR.

\bibitem[{Cer et~al.(2017)Cer, Diab, Agirre, Lopez-Gazpio, and Specia}]{cer-etal-2017-semeval}
Daniel Cer, Mona Diab, Eneko Agirre, I{\~n}igo Lopez-Gazpio, and Lucia Specia. 2017.
\newblock \href {https://doi.org/10.18653/v1/S17-2001} {{S}em{E}val-2017 task 1: Semantic textual similarity multilingual and crosslingual focused evaluation}.
\newblock In \emph{Proceedings of the 11th International Workshop on Semantic Evaluation ({S}em{E}val-2017)}, pages 1--14, Vancouver, Canada. Association for Computational Linguistics.

\bibitem[{Clark et~al.(2022)Clark, De~Las~Casas, Guy, Mensch, Paganini, Hoffmann, Damoc, Hechtman, Cai, Borgeaud, Van Den~Driessche, Rutherford, Hennigan, Johnson, Cassirer, Jones, Buchatskaya, Budden, Sifre, Osindero, Vinyals, Ranzato, Rae, Elsen, Kavukcuoglu, and Simonyan}]{clark2022unified}
Aidan Clark, Diego De~Las~Casas, Aurelia Guy, Arthur Mensch, Michela Paganini, Jordan Hoffmann, Bogdan Damoc, Blake Hechtman, Trevor Cai, Sebastian Borgeaud, George~Bm Van Den~Driessche, Eliza Rutherford, Tom Hennigan, Matthew~J Johnson, Albin Cassirer, Chris Jones, Elena Buchatskaya, David Budden, Laurent Sifre, Simon Osindero, Oriol Vinyals, Marc'Aurelio Ranzato, Jack Rae, Erich Elsen, Koray Kavukcuoglu, and Karen Simonyan. 2022.
\newblock \href {https://proceedings.mlr.press/v162/clark22a.html} {Unified scaling laws for routed language models}.
\newblock In \emph{Proceedings of the 39th International Conference on Machine Learning}, volume 162 of \emph{Proceedings of Machine Learning Research}, pages 4057--4086. PMLR.

\bibitem[{Devlin et~al.(2019)Devlin, Chang, Lee, and Toutanova}]{devlin-etal-2019-bert}
Jacob Devlin, Ming-Wei Chang, Kenton Lee, and Kristina Toutanova. 2019.
\newblock \href {https://doi.org/10.18653/v1/N19-1423} {{BERT}: Pre-training of deep bidirectional transformers for language understanding}.
\newblock In \emph{Proceedings of the 2019 Conference of the North {A}merican Chapter of the Association for Computational Linguistics: Human Language Technologies, Volume 1 (Long and Short Papers)}, pages 4171--4186, Minneapolis, Minnesota. Association for Computational Linguistics.

\bibitem[{Dolan and Brockett(2005)}]{dolan2005automatically}
Bill Dolan and Chris Brockett. 2005.
\newblock \href {https://www.microsoft.com/en-us/research/publication/automatically-constructing-a-corpus-of-sentential-paraphrases/} {Automatically constructing a corpus of sentential paraphrases}.
\newblock In \emph{Third International Workshop on Paraphrasing (IWP2005)}. Asia Federation of Natural Language Processing.

\bibitem[{Geiping and Goldstein(2023)}]{geiping2022cramming}
Jonas Geiping and Tom Goldstein. 2023.
\newblock \href {https://proceedings.mlr.press/v202/geiping23a.html} {Cramming: Training a language model on a single {GPU} in one day.}
\newblock In \emph{Proceedings of the 40th International Conference on Machine Learning}, volume 202 of \emph{Proceedings of Machine Learning Research}, pages 11117--11143. PMLR.

\bibitem[{Hinton et~al.(2015)Hinton, Vinyals, and Dean}]{hinton2015distilling}
Geoffrey Hinton, Oriol Vinyals, and Jeff Dean. 2015.
\newblock \href {http://arxiv.org/abs/1503.02531} {Distilling the knowledge in a neural network}.

\bibitem[{Hoffmann et~al.(2022)Hoffmann, Borgeaud, Mensch, Buchatskaya, Cai, Rutherford, de~Las~Casas, Hendricks, Welbl, Clark, Hennigan, Noland, Millican, van~den Driessche, Damoc, Guy, Osindero, Simonyan, Elsen, Vinyals, Rae, and Sifre}]{hoffmann2022training}
Jordan Hoffmann, Sebastian Borgeaud, Arthur Mensch, Elena Buchatskaya, Trevor Cai, Eliza Rutherford, Diego de~Las~Casas, Lisa~Anne Hendricks, Johannes Welbl, Aidan Clark, Thomas Hennigan, Eric Noland, Katherine Millican, George van~den Driessche, Bogdan Damoc, Aurelia Guy, Simon Osindero, Kar\'{e}n Simonyan, Erich Elsen, Oriol Vinyals, Jack Rae, and Laurent Sifre. 2022.
\newblock \href {https://proceedings.neurips.cc/paper_files/paper/2022/file/c1e2faff6f588870935f114ebe04a3e5-Paper-Conference.pdf} {An empirical analysis of compute-optimal large language model training}.
\newblock In \emph{Advances in Neural Information Processing Systems}, volume~35, pages 30016--30030. Curran Associates, Inc.

\bibitem[{Izsak et~al.(2021)Izsak, Berchansky, and Levy}]{izsak-etal-2021-train}
Peter Izsak, Moshe Berchansky, and Omer Levy. 2021.
\newblock \href {https://doi.org/10.18653/v1/2021.emnlp-main.831} {How to train {BERT} with an academic budget}.
\newblock In \emph{Proceedings of the 2021 Conference on Empirical Methods in Natural Language Processing}, pages 10644--10652, Online and Punta Cana, Dominican Republic. Association for Computational Linguistics.

\bibitem[{Jha et~al.(2023)Jha, Sherborne, Walsh, Groeneveld, Strubell, and Beltagy}]{jha2023train}
Ananya~Harsh Jha, Tom Sherborne, Evan~Pete Walsh, Dirk Groeneveld, Emma Strubell, and Iz~Beltagy. 2023.
\newblock \href {http://arxiv.org/abs/2305.14864} {How to train your (compressed) large language model}.

\bibitem[{Jiao et~al.(2020)Jiao, Yin, Shang, Jiang, Chen, Li, Wang, and Liu}]{jiao-etal-2020-tinybert}
Xiaoqi Jiao, Yichun Yin, Lifeng Shang, Xin Jiang, Xiao Chen, Linlin Li, Fang Wang, and Qun Liu. 2020.
\newblock \href {https://doi.org/10.18653/v1/2020.findings-emnlp.372} {{T}iny{BERT}: Distilling {BERT} for natural language understanding}.
\newblock In \emph{Findings of the Association for Computational Linguistics: EMNLP 2020}, pages 4163--4174, Online. Association for Computational Linguistics.

\bibitem[{Kaddour et~al.(2023)Kaddour, Key, Nawrot, Minervini, and Kusner}]{kaddour2023train}
Jean Kaddour, Oscar Key, Piotr Nawrot, Pasquale Minervini, and Matt~J. Kusner. 2023.
\newblock \href {http://arxiv.org/abs/2307.06440} {No train no gain: Revisiting efficient training algorithms for transformer-based language models}.

\bibitem[{Kaplan et~al.(2020)Kaplan, McCandlish, Henighan, Brown, Chess, Child, Gray, Radford, Wu, and Amodei}]{kaplan2020scaling}
Jared Kaplan, Sam McCandlish, Tom Henighan, Tom~B. Brown, Benjamin Chess, Rewon Child, Scott Gray, Alec Radford, Jeffrey Wu, and Dario Amodei. 2020.
\newblock \href {http://arxiv.org/abs/2001.08361} {Scaling laws for neural language models}.

\bibitem[{Kobayashi(2018)}]{soskkobayashi2018bookcorpus}
Sosuke Kobayashi. 2018.
\newblock Homemade bookcorpus.
\newblock \url{https://github.com/soskek/bookcorpus}.

\bibitem[{Kudugunta et~al.(2023)Kudugunta, Caswell, Zhang, Garcia, Choquette-Choo, Lee, Xin, Kusupati, Stella, Bapna, and Firat}]{kudugunta2023madlad400}
Sneha Kudugunta, Isaac Caswell, Biao Zhang, Xavier Garcia, Christopher~A. Choquette-Choo, Katherine Lee, Derrick Xin, Aditya Kusupati, Romi Stella, Ankur Bapna, and Orhan Firat. 2023.
\newblock \href {http://arxiv.org/abs/2309.04662} {Madlad-400: A multilingual and document-level large audited dataset}.

\bibitem[{Liu et~al.(2021)Liu, Wang, Kasai, Hajishirzi, and Smith}]{liu-etal-2021-probing-across}
Zeyu Liu, Yizhong Wang, Jungo Kasai, Hannaneh Hajishirzi, and Noah~A. Smith. 2021.
\newblock \href {https://doi.org/10.18653/v1/2021.findings-emnlp.71} {Probing across time: What does {R}o{BERT}a know and when?}
\newblock In \emph{Findings of the Association for Computational Linguistics: EMNLP 2021}, pages 820--842, Punta Cana, Dominican Republic. Association for Computational Linguistics.

\bibitem[{Lu et~al.(2022)Lu, Zhang, Chu, Chen, Zhou, Wu, Chen, and Yang}]{lu2022knowledge}
Chengqiang Lu, Jianwei Zhang, Yunfei Chu, Zhengyu Chen, Jingren Zhou, Fei Wu, Haiqing Chen, and Hongxia Yang. 2022.
\newblock \href {http://arxiv.org/abs/2206.14366} {Knowledge distillation of transformer-based language models revisited}.

\bibitem[{Rajpurkar et~al.(2016)Rajpurkar, Zhang, Lopyrev, and Liang}]{rajpurkar-etal-2016-squad}
Pranav Rajpurkar, Jian Zhang, Konstantin Lopyrev, and Percy Liang. 2016.
\newblock \href {https://doi.org/10.18653/v1/D16-1264} {{SQ}u{AD}: 100,000+ questions for machine comprehension of text}.
\newblock In \emph{Proceedings of the 2016 Conference on Empirical Methods in Natural Language Processing}, pages 2383--2392, Austin, Texas. Association for Computational Linguistics.

\bibitem[{Sanh et~al.(2020)Sanh, Debut, Chaumond, and Wolf}]{sanh2020distilbert}
Victor Sanh, Lysandre Debut, Julien Chaumond, and Thomas Wolf. 2020.
\newblock \href {http://arxiv.org/abs/1910.01108} {Distilbert, a distilled version of bert: smaller, faster, cheaper and lighter}.

\bibitem[{Socher et~al.(2013)Socher, Perelygin, Wu, Chuang, Manning, Ng, and Potts}]{socher-etal-2013-recursive}
Richard Socher, Alex Perelygin, Jean Wu, Jason Chuang, Christopher~D. Manning, Andrew Ng, and Christopher Potts. 2013.
\newblock \href {https://aclanthology.org/D13-1170} {Recursive deep models for semantic compositionality over a sentiment treebank}.
\newblock In \emph{Proceedings of the 2013 Conference on Empirical Methods in Natural Language Processing}, pages 1631--1642, Seattle, Washington, USA. Association for Computational Linguistics.

\bibitem[{Sun et~al.(2020)Sun, Yu, Song, Liu, Yang, and Zhou}]{sun-etal-2020-mobilebert}
Zhiqing Sun, Hongkun Yu, Xiaodan Song, Renjie Liu, Yiming Yang, and Denny Zhou. 2020.
\newblock \href {https://doi.org/10.18653/v1/2020.acl-main.195} {{M}obile{BERT}: a compact task-agnostic {BERT} for resource-limited devices}.
\newblock In \emph{Proceedings of the 58th Annual Meeting of the Association for Computational Linguistics}, pages 2158--2170, Online. Association for Computational Linguistics.

\bibitem[{Turc et~al.(2019)Turc, Chang, Lee, and Toutanova}]{turc2019wellread}
Iulia Turc, Ming-Wei Chang, Kenton Lee, and Kristina Toutanova. 2019.
\newblock \href {http://arxiv.org/abs/1908.08962} {Well-read students learn better: On the importance of pre-training compact models}.

\bibitem[{Wang et~al.(2018)Wang, Singh, Michael, Hill, Levy, and Bowman}]{wang-etal-2018-glue}
Alex Wang, Amanpreet Singh, Julian Michael, Felix Hill, Omer Levy, and Samuel Bowman. 2018.
\newblock \href {https://doi.org/10.18653/v1/W18-5446} {{GLUE}: A multi-task benchmark and analysis platform for natural language understanding}.
\newblock In \emph{Proceedings of the 2018 {EMNLP} Workshop {B}lackbox{NLP}: Analyzing and Interpreting Neural Networks for {NLP}}, pages 353--355, Brussels, Belgium. Association for Computational Linguistics.

\bibitem[{Wang et~al.(2020)Wang, Wei, Dong, Bao, Yang, and Zhou}]{wang2020minilm}
Wenhui Wang, Furu Wei, Li~Dong, Hangbo Bao, Nan Yang, and Ming Zhou. 2020.
\newblock Minilm: deep self-attention distillation for task-agnostic compression of pre-trained transformers.
\newblock In \emph{Proceedings of the 34th International Conference on Neural Information Processing Systems}, NIPS'20, Red Hook, NY, USA. Curran Associates Inc.

\bibitem[{Wang et~al.(2023)Wang, Weissweiler, Sch{\"u}tze, and Plank}]{wang-etal-2023-distill}
Xinpeng Wang, Leonie Weissweiler, Hinrich Sch{\"u}tze, and Barbara Plank. 2023.
\newblock \href {https://doi.org/10.18653/v1/2023.acl-short.157} {How to distill your {BERT}: An empirical study on the impact of weight initialisation and distillation objectives}.
\newblock In \emph{Proceedings of the 61st Annual Meeting of the Association for Computational Linguistics (Volume 2: Short Papers)}, pages 1843--1852, Toronto, Canada. Association for Computational Linguistics.

\bibitem[{Warstadt et~al.(2019)Warstadt, Singh, and Bowman}]{warstadt-etal-2019-neural}
Alex Warstadt, Amanpreet Singh, and Samuel~R. Bowman. 2019.
\newblock \href {https://doi.org/10.1162/tacl_a_00290} {Neural network acceptability judgments}.
\newblock \emph{Transactions of the Association for Computational Linguistics}, 7:625--641.

\bibitem[{Williams et~al.(2018)Williams, Nangia, and Bowman}]{williams-etal-2018-broad}
Adina Williams, Nikita Nangia, and Samuel Bowman. 2018.
\newblock \href {https://doi.org/10.18653/v1/N18-1101} {A broad-coverage challenge corpus for sentence understanding through inference}.
\newblock In \emph{Proceedings of the 2018 Conference of the North {A}merican Chapter of the Association for Computational Linguistics: Human Language Technologies, Volume 1 (Long Papers)}, pages 1112--1122, New Orleans, Louisiana. Association for Computational Linguistics.

\bibitem[{Zhu et~al.(2015)Zhu, Kiros, Zemel, Salakhutdinov, Urtasun, Torralba, and Fidler}]{zhu2015aligning}
Yukun Zhu, Ryan Kiros, Rich Zemel, Ruslan Salakhutdinov, Raquel Urtasun, Antonio Torralba, and Sanja Fidler. 2015.
\newblock \href {https://doi.org/10.1109/ICCV.2015.11} {Aligning books and movies: Towards story-like visual explanations by watching movies and reading books}.
\newblock In \emph{2015 IEEE International Conference on Computer Vision (ICCV)}, pages 19--27.

\end{thebibliography}
\bibliographystyle{acl_natbib}

\appendix

\section{Appendix}

\subsection{Implementation Details}

We use the code base from \citet{izsak-etal-2021-train} for both pretraining and finetuning of \nokd models. In the case of \kd models, we utilize the code base introduced by \citet{wang-etal-2023-distill}, which itself builds upon the work of \citet{izsak-etal-2021-train}. Our code is available at \url{https://github.com/MinhDucBui/revisiting_distillation}.

\subsection{Pretraining Details} \label{sec:pretraining}

Our pretraining pipeline employs a batch size of 1024, employing gradient accumulation with a batch size of 256. We adopt a time-based learning rate schedule with a linear curve. The peak learning rate is set to 5e-4 for distillation strategies and 1e-3 for \nokd. We opt for a warmup proportion of 0.06 for both scenarios. Utilizing the AdamW optimizer with $(\beta_1, \beta_2) = (0.9, 0.98)$ and $\epsilon = 1e-6$, we conduct training with mixed precision techniques. 

We measure compute budget by wall-clock time. All experiments are conducted on NVIDIA A100. Training our 6-layer model for a single epoch requires around 4 hours of wall-clock training time, while the 12-layer model demands approximately 11 hours. Scaling up the 6-layer model to 27.9B tokens extends the training duration to about 24 hours. Fine-tuning on GLUE with a single A100 GPU, coupled with grid-hyperparameter search, consumes up to 50 hours for the 6-layer models and nearly 100 hours for the 12-layer variants.

\subsection{GLUE Details} \label{sec:glue}

We provide a brief overview of each dataset within GLUE. For additional information regarding each data split, evaluation metric and more, see \citet{wang-etal-2018-glue}.

\paragraph{CoLA} The Corpus of Linguistic Acceptability \cite{warstadt-etal-2019-neural} comprises English acceptability judgments sourced from books and journal articles on linguistic theory.

\paragraph{SST-2} The Stanford Sentiment Treebank \cite{socher-etal-2013-recursive} includes sentences extracted from movie reviews, along with human annotations of their binary sentiment.

\paragraph{MRPC} The Microsoft Research Paraphrase Corpus \cite{dolan2005automatically} consists of sentence pairs automatically extracted from online news sources, with human annotations indicating whether the sentences are semantically equivalent.

\paragraph{QQP} The Quora Question Pairs dataset is a compilation of question pairs from the community question-answering website. The objective is to determine whether a pair of questions are semantically equivalent.

\paragraph{STS-B} The Semantic Textual Similarity Benchmark \cite{cer-etal-2017-semeval} contains sentence pairs with a human annotated similarity score ranging from 1 to 5.

\paragraph{MNLI} The Multi-Genre Natural Language Inference Corpus \cite{williams-etal-2018-broad} is a crowd-sourced collection of sentence pairs with textual entailment annotations. The task involves predicting whether a premise sentence entails, contradicts, or is neutral with respect to a hypothesis.

\paragraph{QNLI} The Stanford Question Answering Dataset \cite{rajpurkar-etal-2016-squad} is a question-answering dataset comprising question-paragraph pairs, with the task of determining whether the context sentence contains the answer to the question.

\paragraph{RTE} The Recognizing Textual Entailment (RTE) datasets originate from annual textual entailment challenges. The dataset is standardized to a two-class split, collapsing neutral and contradiction into "not entailment" for consistency.

%\subsection{Model Details} \label{sec:model-details}

%We report our hyperparameters used in Table \ref{}.
%\begin{table}[]
%    \centering
%    \begin{tabular}{c|c|c}
%         Hyperparameter & %Model\textsubscript{6} & %Model\textsubscript{12} \\
%         \hline
%         Number of Layers & 6 & 12 \\
%         Hidden Size & 768 & 1024 \\
%         FFN Inner Hidden Size & 3072 & 4096 \\
%         Attention Heads & 12 & 16 \\
%         Total Parameters & 12 & 16 \\
%    \end{tabular}
%    \caption{Hyperparameters employed in both our 6-layer (Model\textsubscript{6}) and 12-layer (Model\textsubscript{12}) models, encompassing both \nokd and \kd configurations.}
%    \label{tab:my_label}
%\end{table}

\end{document}